\documentclass[runningheads]{llncs}
\usepackage[T1]{fontenc}
\usepackage{graphicx,verbatim}
\usepackage{graphicx}%
\usepackage{multirow}%
\usepackage{amsmath,amssymb,amsfonts}%
\usepackage{mathrsfs}%
\usepackage[title]{appendix}%
\usepackage{xcolor}%
\usepackage{textcomp}%
\usepackage{manyfoot}%
\usepackage{booktabs}%
\usepackage{algorithm}%
\usepackage{algorithmicx}%
\usepackage{algpseudocode}%
\usepackage{listings}%
\usepackage{graphicx}
\usepackage{wrapfig}
\usepackage{mathrsfs}
\usepackage{amssymb}
\usepackage[mathscr]{euscript}
\usepackage{caption}
\usepackage{hyperref}  
\begin{document}
%
\title{SCOPE: Speech-guided COllaborative PErception Framework for Surgical Scene Segmentation}
\titlerunning{SCOPE}
%

\author{Jecia Z.Y. Mao, Francis X Creighton, Russell H Taylor, Manish Sahu}  
\authorrunning{Mao et al.}
\institute{LCSR, Johns Hopkins University \\
    \email{zmao16@jh.edu, francis.creighton@jhmi.edu, rht@jhu.edu, manish.sahu@jhu.edu}}

\maketitle              
%


\begin{abstract}

Accurate segmentation and tracking of relevant elements of the surgical scene is crucial to enable context-aware intraoperative assistance and decision making.
Current solutions remain tethered to domain-specific, supervised models that rely on labeled data and required domain-specific data to adapt to new surgical scenarios and beyond predefined label categories. 
Recent advances in prompt-driven vision foundation models (VFM) have enabled open-set, zero-shot segmentation across heterogeneous medical images. However, dependence of these models on manual visual or textual cues restricts their deployment in introperative surgical settings.
We introduce a speech-guided collaborative perception (SCOPE) framework that integrates reasoning capabilities of large language model (LLM) with perception capabilities of open-set VFMs to support on-the-fly segmentation, labeling and tracking of surgical instruments and anatomy in intraoperative video streams.
A key component of this framework is a collaborative perception agent, which generates top candidates of VFM-generated segmentation and incorporates intuitive speech feedback from clinicians to guide the segmentation of surgical instruments in a natural human-machine collaboration paradigm. Afterwards, instruments themselves serve as interactive pointers to label additional elements of the surgical scene. 
We evaluated our proposed framework on a subset of publicly available Cataract1k dataset and an in-house ex-vivo skull-base dataset to demonstrate its potential to generate on-the-fly segmentation and tracking of surgical scene. Furthermore, we demonstrate its dynamic capabilities through a live mock ex-vivo experiment. 
This human-AI collaboration paradigm showcase the potential of developing adaptable, hands-free, surgeon-centric tools for dynamic operating-room environments.


\end{abstract}
%
\section{Introduction}

Surgical scene segmentation in endoscopic video is a fundamental task in computer-assisted surgery, enabling downstream tasks such as real-time intraoperative guidance, context-aware assistance, and automated postoperative analysis~\cite{vercauteren2019cai4cai}.
Traditional approaches have predominantly relied on deep learning models trained in a supervised fashion on large, manually annotated datasets~\cite{ahmed2024deep}. 
While these methods have achieved high accuracy, they struggle to generalize beyond domain-specific data and labels, which restricts their utility in intraoperative settings, where responsiveness to unseen conditions is necessary.

Recent progress in vision foundation models (VFMs) has introduced new possibilities for open-set, prompt-driven segmentation. 
Models like the Segment Anything Model (SAM)~\cite{kirillov2023SAM} enable category-agnostic segmentation using visual prompts such as clicks or bounding boxes. Grounded SAM (GSAM)~\cite{ren2024GSAM} extends this paradigm to text-promptable segmentation by integrating open-vocabulary object detection model Grounding DINO~\cite{liu2024grounding} to enable text-prompted segmentation across a broad label space. 
Emerging research in reasoning segmentation models, such as LISA++~\cite{yang2023lisa++}, explore the use of implicit text queries to support semantically meaningful and context-aware segmentation through multimodal models that combine visual perception with natural language understanding. 
Together, these advances mark a shift toward more generalizable and interactive systems.
However, current interactive methods remain limited by their dependence on manual visual or textual inputs, such as mouse clicks or keyboard-entered queries~\cite{soberanis2024gsam_cutie}. While effective for dataset annotation, these interfaces are poorly suited for sterile intraoperative environments.
For such systems to be usable in practice, they must fit seamlessly into the surgical workflow and support intuitive, hands-free interaction.

In this work, we introduce a speech-guided collaborative perception (SCOPE) framework tailored for intraoperative use. Our system combines the natural language understanding and reasoning capabilities of large language models (LLMs) with the visual perception of open-set VFMs to enable hands-free, on-the-fly segmentation and tracking of surgical instruments and anatomical structures. At the core of the framework lies a speech-based perception agent that collaborates with the clinician through spoken commands to refine segmentation outputs, label scene elements, and adapt dynamically to evolving surgical contexts.

We evaluate our framework on a publicly available cataract dataset and an in-house ex-vivo microscopic skull-base dataset. Across both datasets, our system demonstrates high performance while offering significant advantages in usability and adaptability. These results highlight the potential of speech-guided, collaborative AI systems for augmenting intraoperative context awareness and decision-making in complex, dynamic surgical environments.

\section{Related Work}

\textbf{Promptable image segmentation}
Prompt-driven VFMs have significantly advanced generalizable segmentation in medical imaging~\cite{zhang2023survey,zhou2024survey}.
In the surgical domain, a recent benchmarking study~\cite{soberanis2024gsam_cutie} demonstrated superior performance of text-promotable GSAM with Cutie for surgical video annotation task.
TPSIS~\cite{zhou2023TPSIS} adopted a reasoning segmentation approach~\cite{yang2023lisa++} approach to addresses class-specific surgical instrument differentiation using transformer-based attention and auxiliary depth cues but remains constrained by its reliance on labeled surgical datasets. 
Similarly, RSVIS~\cite{wang2024RSVIS} supports referring video segmentation with temporal reasoning across robotic surgery frames, yet it depends on extensive domain adaptation and vocabulary grounding. 
Rather than fine-tuning on surgery-specific data, which often require massive computational resources and risk overfitting to narrow distributions, we adopted zero-shot VFMs for our framework as we prioritize generalization across procedures and ease of deployment in data-constrained surgical environments.

\noindent \textbf{Interactive visual reasoning}
Frameworks like Visual ChatGPT~\cite{wu2023visualChatGPT} demonstrate how LLMs can orchestrate vision modules via multimodal prompts, enabling iterative visual reasoning through dialogue. 
Inspired by this paradigm, our work leverages large language models not merely for communication, but for orchestrating a flexible, speech-driven visual reasoning pipeline. Unlike Visual ChatGPT, which focuses on general visual tasks with static images, our system is designed for dynamic introperative surgical video streams. We emphasize hands-free interaction, real-time adaptability and goal-cnetric collaboration.

\section{Method}
\begin{figure}[t]
  \makebox[\textwidth][c]{%
    \includegraphics[width=1.0\linewidth]{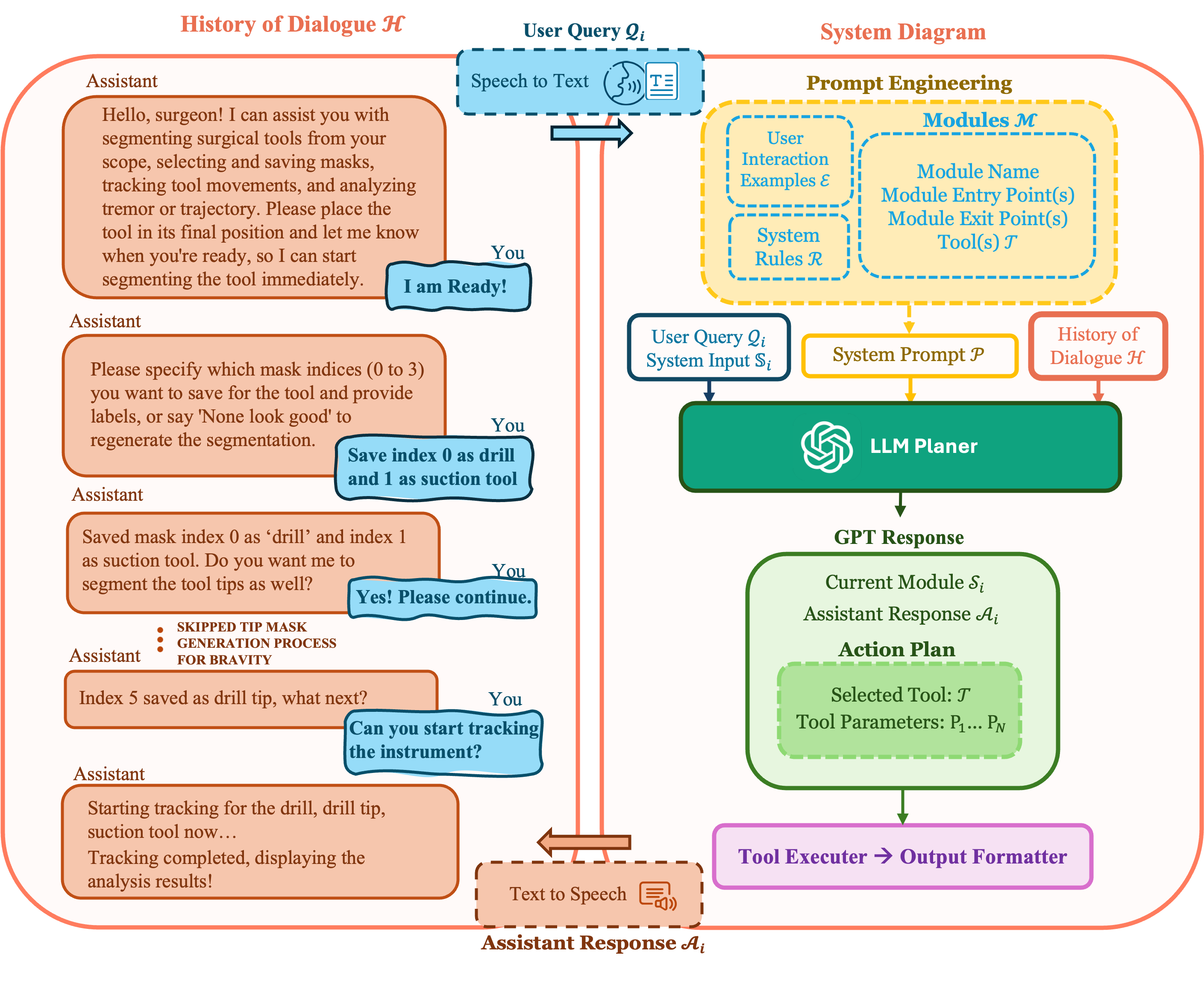}%
  }
  \caption{System diagram. Left: tentative user interactions. Right: underlying components that enables such structured workflow.}
  \label{fig:sys-diag}
\end{figure}
\subsection{System Architecture}
Our interactive framework is designed to support natural human-AI collaboration through structured verbal interaction.
It uses a cloud-based LLM (GPT-4.1 mini) agent for stepwise reasoning and generate action responses for VFMs to allow users to perform visual tasks through interactive dialogue. 
The speech AI agent generates responses in the form:
\[\mathcal{R}_i =\{\text{Action},\text{Text Response}\} = LLM(\mathcal{Q}_i, \mathbb{S}_i,\mathcal{P},\mathcal{H}_{(<i)})\]
where $\mathcal{Q}_i$ denotes the  current user query, $\mathbb{S}_i$ the system inputs, $\mathcal{P}$ the system prompt, and $\mathcal{H}$ the history of prior interactions. 
The response $\mathcal{R}_i$ guides both the internal function calls to the underlying VFMs and verbal responses to the user.
To enable an intuitive and straightforward human-in-the-loop verbal interaction, we created a system prompt $\mathcal{P}$ that defines the system description, constrains and the workflow: 
\[\mathcal{P}= \{\mathcal{M}_i, \mathcal{T}_j, \mathcal{E}, \mathcal{R}\}\] \textbf{Module} $\mathcal{M}$ represent the key phases of interaction to ensure the workflow progresses in a meaningful and context-aware manner.
Each module is associated with a specific prompt $\mathscr{p}$ that defines its entry/exit criteria, allowed tool calls, and in-context examples for user-agent interactions.
\[\mathcal{M} = \{InteractiveMode, Segmentation, SelectMask, Tracking\}\]
\noindent \textbf{Tools} \textbf{$\mathcal{T}$} define functional VFM modules invoked at each state. For example, the \textit{segment} tool employs Grounding DINO and SAM to generate segmentation masks from spoken input, while \textit{display} tools assist in visualizing candidates for user selection. Modules guide the sequence of actions, and tools implement the underlying visual computation.
\noindent \textbf{In-context examples} \textbf{$\mathcal{E}$} define demonstrations of task-specific queries and expected system behavior to facilitate few-shot generalization by the LLM. These examples guide the agent across states and selecting the appropriate tools.
\noindent \textbf{Rules} \textbf{$\mathcal{R}$} define the structured interactions of the framework. This includes explicit formatting of the output (\textit{JSON} syntax) to enable functional calls. 
\subsection{Interactive Workflow}
The system start with the \textit{InteractiveMode} state with a greeting. We use a lavalier microphone, attached to the apron, to stream real-time audio response from the user. The audio input is then transcribed it using OpenAI Whisper. 
E.g. given the spoken query (e.g., “segment the surgical instruments”), agent transition into the \textit{Segment} state with "surgical instruments" as the text query.
\begin{figure}
    \centering
    \includegraphics[width=1 \linewidth]{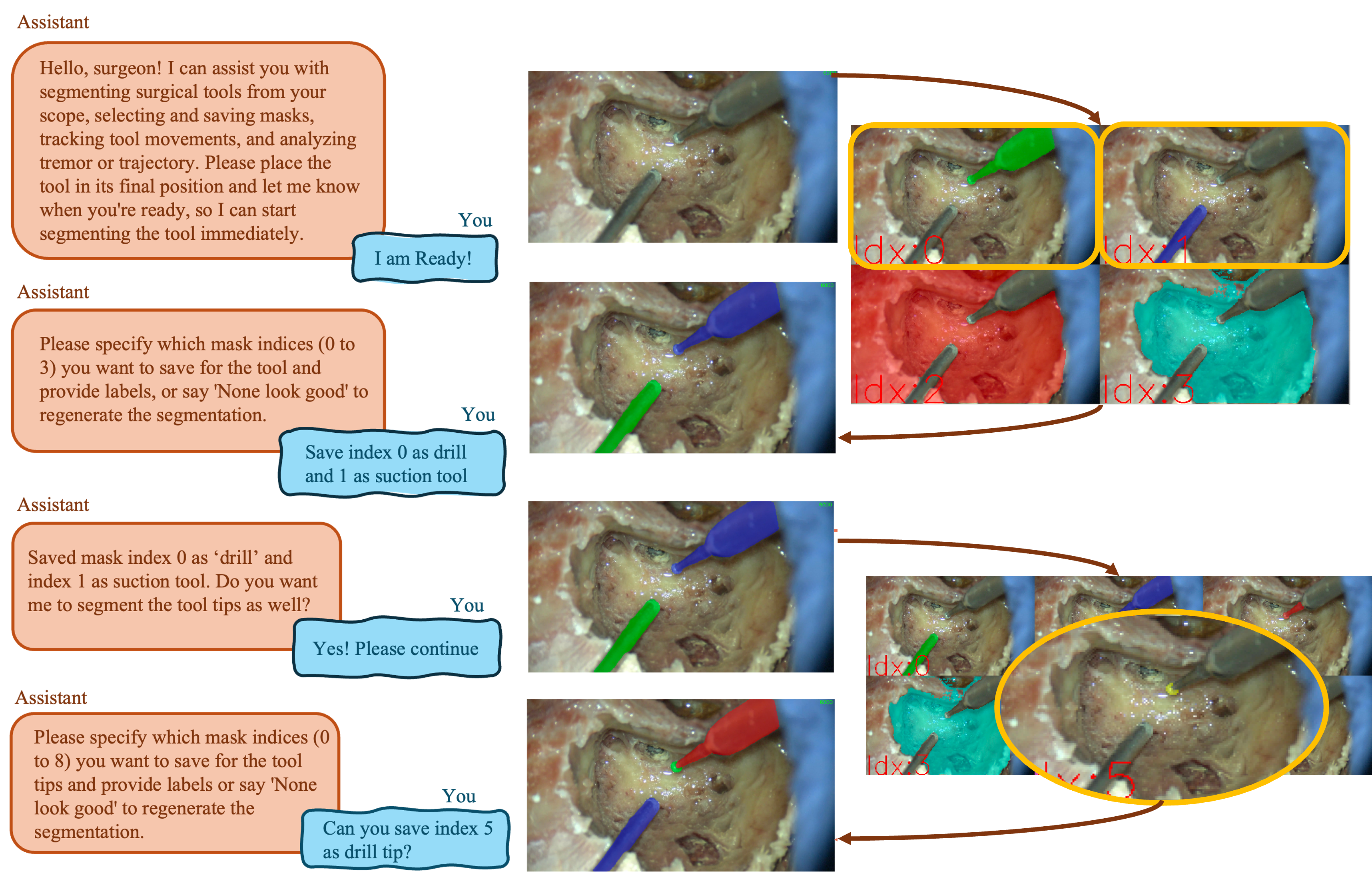}
    \caption{Segmentation workflow: a stage-by-stage visualization of system output while delineating each instrument and its tip.}
    \label{fig:enter-label}
\end{figure}

\noindent\textbf{Instrument Segmentation and Tracking}
Given a text query (e.g., “surgical instruments”), the query undergoes an expansion process which augments the initial query into semantically similar alternatives (e.g., ``surgical tools'', ``gray instruments'') to increase model recall. 
The spoken command is expanded into semantically related prompts that drive GSAM and LISA++ to generate candidate masks. After applying \textbf{ranking heuristics} and deduplication, the six highest-scoring, non-overlapping masks are displayed. This approach eliminates the possibilities of obtaining overly crowded grid for users to select. If none is satisfactory, the system advances to the next $\textbf{display iteration}$. 
The surgeon verbally selects a mask and assigns its label, completing the segmentation step.
Once instruments are segmented, we use visual promptable VFM for video object segmentation (SAM2 or Cutie) for propagating the mask across video frames. To support tip-based tracking, the system guides the user through another round of segmentation stage to identifying the tip of the instrument. The intersection of the mask boundaries of tip and shaft defines a consistent and geometrically stable tip landmark. For instruments without a distinguishable tip (e.g., suction devices), we instead extract a point along the medial axis of the shaft region, which remains stable due to its rigid geometry. The tip point is tracked by extracting the boundary point along the instrument’s principal axis (computed via principal component analysis). We found that this approach generates tracking point that remains consistent across frames without invoking a separate VFM for point tracking.
\subsubsection{Anatomy Segmentation via Virtual Cursor} 
We enable hands-free anatomy segmentation with a \emph{virtual cursor} tied to the instrument tip. The cursor’s position—offset along the tool’s principal axis—is tracked in every frame. A monocular depth VFM (DepthAnything~\cite{yang2024depthanything}) estimates per-pixel depth; when the tip region contains enough pixels within a preset depth band, a “click’’ is inferred. The cursor location is then fed to SAM as a positive point prompt, yielding an anatomy mask that is propagated over time by the video-segmentation model, so instruments and contacted tissue are segmented concurrently.
\begin{figure}[H]
\centering\includegraphics[width=1\linewidth]{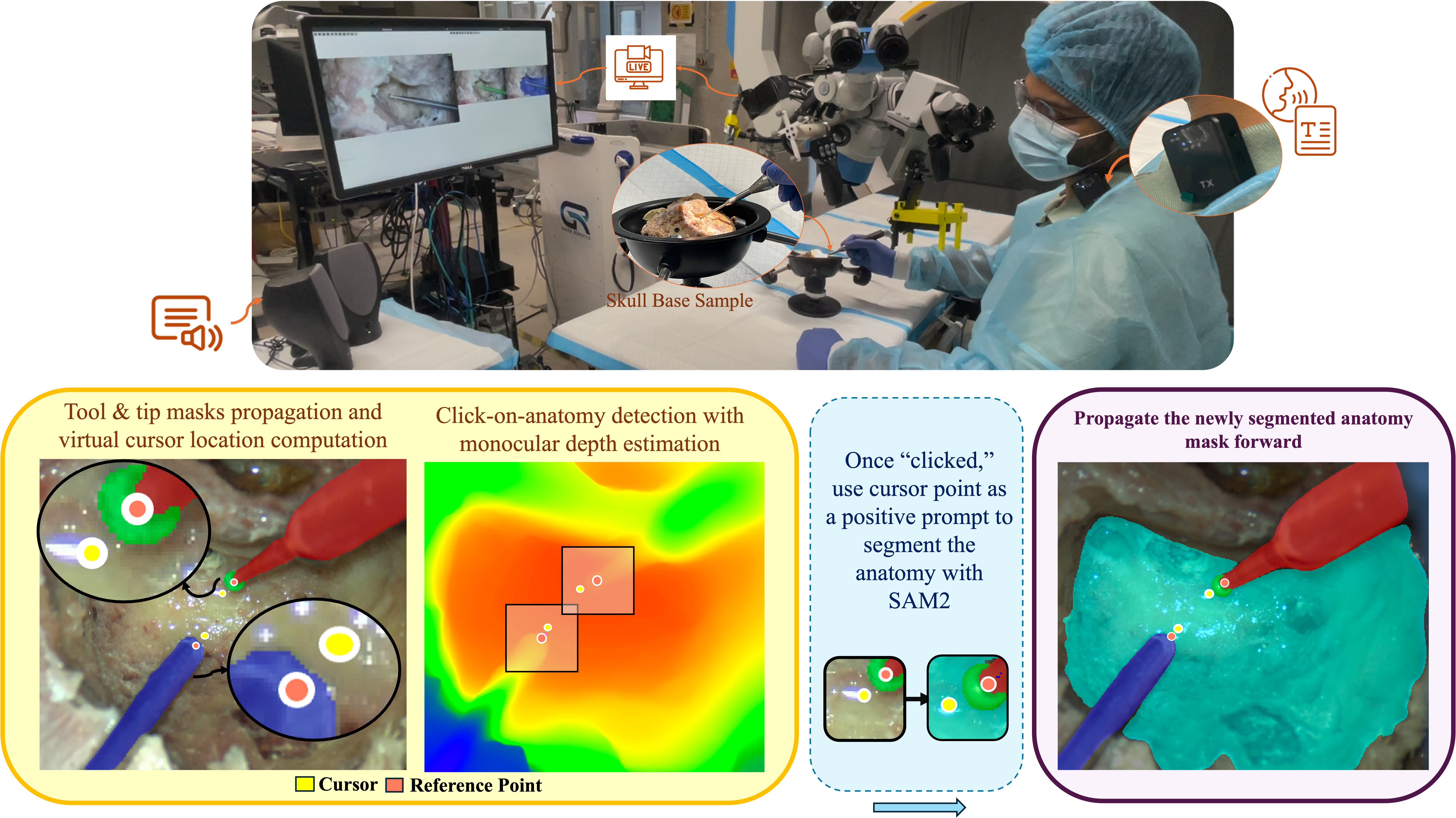}
    \caption{Top: system setup during real-time case study. Bottom: virtual cursor workflow that helps segmenting the anatomy via surface contact.}
    \label{fig:enter-label}
\end{figure}

\section{Experiments and Results}
We benchmarked our two core modules -- interactive segmentation and video tracking -- on (i) a subset of ten videos of publicly available Cataract1k dataset 
~\cite{ghamsarian2023cataract1k} and (ii) an in-house ex-vivo skull-base dataset containing five videos. Segmentation accuracy is reported with Dice score (DSC) and average surface distance (ASD); tracking quality uses their frame-wise means (mDSC, mASD).
\begin{table}
  \begin{minipage}{0.45\textwidth}
  \centering
  \caption{Comparison on Initial Segmentation}
  \label{tab:sota_segmentation}
  \begin{tabular}{@{}llcccc@{}}
    \toprule
    Anatomy & Method  & DSC & ASD & \#Iter. & Time(sec) \\
    \midrule
    \multirow{2}{*}{Eye}
      & LISA++          & 0.46 & 26.53 & 1.0 & 4.44 \\
      & GSAM   & 0.82 & 2.83 & 1.3 & 1.16 \\
    \midrule
    \multirow{2}{*}{Skull Base}
      & LISA++          & 0.74 & 27.52 & 1.0 & 4.44 \\
      & GSAM   & 0.93 & 5.52 & 1.0 & 1.28\\
    \bottomrule
  \end{tabular}
  \end{minipage}
  \hfill
  \begin{minipage}{0.45\textwidth}
  \centering
  \caption{Comparison on Mask Propagation}
  \label{tab:sota_tracking}
  \begin{tabular}{@{}llcc@{}}
    \toprule
    Anatomy & Method  & mDSC & mASD \\
    \midrule
    \multirow{2}{*}{Eye}
      & CUTIE  & 0.840 & 2.736 \\
      & SAM2   & 0.818 & 3.861\\
    \midrule
    \multirow{2}{*}{Skull Base}
      & CUTIE  & 0.973 & 2.54 \\
      & SAM2   & 0.941 & 5.558 \\
    \bottomrule
  \end{tabular}
  \end{minipage}
\textit{Notes.} \#Iter.: average number of display iterations until correct mask; Time: runtime per iteration in seconds.
\end{table}

We compared explicit-query GSAM with reasoning-based LISA++ for interactive surgical-tool segmentation. LISA++ prompts were enriched with tool location, appearance, and tissue-interaction cues to encourage instance separation, yet the model still merged multiple instruments in complex cataract scenes. GSAM performed consistently better: with our ranking heuristic it returned the correct mask in the first iteration for 8 / 10 cataract videos and all 5 skull-base videos, whereas LISA++ was only modestly reliable in simpler one or two-tool skull-base cases (Table \ref{tab:sota_segmentation}).

For video segmentation and tracking, CUTIE’s built-in temporal memory lets it grow the mask as more of the instrument appears, so when a shaft is only partially visible at first and later comes fully into view, the entire tool is automatically captured. In the same scenario SAM2 keeps echoing the original, partial mask, giving CUTIE a clear advantage whenever objects gradually reveal themselves or re-emerge after occlusion.
\subsubsection{Case study on live video stream} To assess the  usability of our framework in a realistic setting, we conducted a mock surgical experiment (see Figure~\ref{fig:enter-label}) on a live surgical video stream. Speech commands, captured by a lavalier mic, prompted Grounded SAM to segment the instruments, while CUTIE handled frame-wise tracking. The correct segmentation mask for surgical instrument is retrieved in the first round; a follow-up isolated its tip, also within the first iteration. Both tool tip and shaft were then stably tracked. When the tip contacted tissue, the system automatically launched anatomy segmentation, confirming seamless end-to-end interactions. For this reason, CUTIE would be more robust in application on surgical scenes.

\section{Discussion and Conclusion}
We present a speech-guided perception system for intraoperative videos that lets surgeons issue natural, hands-free commands to segment and track instruments and anatomy on the fly.
Evaluation on two endoscopic datasets and a live mock experiment confirm real-time feasibility.
Despite these promising results, the framework has certain limitations. (i) latency of the system response time must be reduced for more efficient interactions; and (ii) our study only a mock procedure, not the full complexity of operating-room workflows.
Future work will include evaluating longer, varied cases (laparoscopic, robotic, open), accelerating inference via on-device deployment of originally cloud-based models, and further evaluation to statistically validate the system with surgeon-in-the-loop. These efforts will move the framework from prototype to a deployable assistant that enhances intraoperative context awareness and decision-making.

\section{Supplementary Material}
We will release our code and system prompt at 
\href{https://github.com/LCSR-CIIS/SCOPE}{\nolinkurl{https://github.com/LCSR-CIIS/SCOPE}}.

\bibliography{bibliography}
\bibliographystyle{splncs04}
\end{document}